\documentclass[runningheads]{llncs}

\usepackage{graphicx}
\usepackage{graphics}
\usepackage{comment}
\usepackage{amsmath,amssymb}
\usepackage{color}
\usepackage{url}
\usepackage{hyperref}

\usepackage{array}
\usepackage{multirow}
\usepackage[ruled,vlined]{algorithm2e}

\newif\ifreview
\reviewfalse

\ifreview
	\usepackage{lineno}

	\linenumbers
\fi

\begin{document}


\def\SubNumber{110}

\def\GCPRTrack{Regular Track}

\title{Event-Based Feature Tracking in Continuous Time with Sliding Window Optimization}

\ifreview
	\titlerunning{DAGM GCPR 2021 Submission \SubNumber{}. CONFIDENTIAL REVIEW COPY.}
	\authorrunning{DAGM GCPR 2021 Submission \SubNumber{}. CONFIDENTIAL REVIEW COPY.}
	\author{DAGM GCPR 2021 - \GCPRTrack{}}
	\institute{Paper ID \SubNumber}
\else
	\titlerunning{Event-Based Feature Tracking in Continuous Time with SWO}

	\author{Jason Chui\orcidID{0000-0001-8188-9734} \and
	Simon Klenk \and \\
	Daniel Cremers\orcidID{0000-0002-3079-7984}}
	
	\authorrunning{J. Chui et al.}
	
	\institute{Technical University of Munich, Germany \\
	\email{\{jason.chui, simon.klenk, cremers\}tum.de}}
\fi

\maketitle              

\begin{abstract}
 We propose a novel method for continuous-time feature tracking in event cameras.  To this end, we track features by aligning events along an estimated trajectory in space-time such that the projection on the image plane results in maximally sharp event patch images. The trajectory is parameterized by $n^{th}$ order B-splines, which are continuous up to $(n-2)^{th}$ derivative. In contrast to previous work, we optimize the curve parameters in a sliding window fashion. On a public dataset we experimentally confirm that the proposed sliding-window B-spline optimization leads to longer and more accurate feature tracks than in previous work.

\end{abstract}

\textbf{Please add the paper submission id and the track name to the paper.}

\section{Introduction}
Event cameras (dynamic vision sensors) are imaging devices which asynchronously measure per-pixel  brightness changes. These sensors are suited for robotics and virtual reality applications, since they offer lower latency,  lower power consumption as well as higher dynamic range and higher temporal resolution compared to frame-based cameras. In order to actually tap into these benefits, computer vision algorithms for event-based sensors need to be developed. However, since event sensors are based on fundamentally different measurement principles than standard frame-based cameras, traditional computer vision algorithms cannot simply be applied to event data, but rather need to be developed from scratch.

Event cameras report per-pixel brightness changes of the observed logarithmic brightness. Each event $\mathbf{e}_i = \{ t_i, \mathbf{x}_i, p_i\}$ is a tuple of a micro-resolution timestamp $t_i$, image plane coordinates $\mathbf{x}_i = (x_i, y_i)$ and the respective polarity change $p_i \in \{-1, 1\}$. The data stream is asynchronous and sparse because an event is transmitted only if the logarithmic brightness changes by a predefined, usually unknown threshold. This is in contrast to frame-based cameras, where each pixel is illuminated during a shared exposure time interval, resulting in an absolute brightness measurement at a fixed frequency.

A common approach is to accumulate events over a fixed time interval into a frame-like structure and apply traditional computer vision methods on them. The drawback of a naive accumulation is that moving edges in the scene result in blurred edges in the image plane. One popular way to correct for this error is by estimating constant optical flow in a certain space-time window, i.e. at a predefined patch location and during a time interval. Each event in the window is then warped to a common reference time using the estimated optical flow. The goal is to create maximally sharp event patch image. This approach is known as motion-compensation \cite{gallego2018unifying}. It is applied to a variety of event-based vision tasks, such as feature tracking \cite{Gehrig19ijcv}, \cite{Seok2020WACV}, ego-motion estimation \cite{Zhu_2019_CVPRUnsupervised}, \cite{ye2018unsupervised} or motion segmentation \cite{stoffregen2019event}.

In case of feature tracking, there remains one main drawback with motion-compensation: Usually, the estimated optical flow is assumed to be constant in a certain (short) time interval, hence the trajectory of events in the x-y-t space-time volume is piece-wise linear. To resolve this problem, we propose to track features with a continuous curve and optimize the curve in the manner of sliding window optimization. In this paper, we propose to employ B-splines as a curve representation in sliding window optimization for two reasons: (1) To account for feature tracks of variable length, we can build n-th order continuous trajectories by adding any number of knots to the curve. (2)  Adding knots or changing knot values only changes the curve {\em locally}, so we can reduce the problem size by setting old knots which are out of scope to a constant value. Our approach is compared to a state-of-the-art event feature tracking algorithm and shows significant improvements in terms of error and feature age. The contributions of this paper can be summarized as follows:

\begin{enumerate}\setlength\itemsep{2mm}
  \item We introduce the first event feature tracking algorithm that uses continuous B-spline functions and employs SE2 warping of events.
  \item We optimize the B-spline parameters of the trajectory in a sliding-window manner.
  \item We experimentally confirm significant improvements in tracking precision and feature age over existing event tracking algorithms.
\end{enumerate}

\section{Related Work}
One line of work performs hybrid tracking by combining frames with events. The advantage of hybrid approaches is that during certain degenerate motions, such as movements parallel to an intensity edge (when no events are triggered),  the frames can still provide useful intensity information for tracking. The work by Gehrig et al. \cite{Gehrig19ijcv} detects features in standard frames and tracks those using events. Their approach employs an event generation model, which is based on frames and estimated optical flow to predict the observed events. To model larger variations in appearance, the feature patches are additionally parameterized by a rigid warp function in the image plane. This method achieves accurate results on a variety of datasets. However, it relies on specialized hardware, such as the Dynamic and Active-pixel Vision Sensor  \cite{brandli2014240}, which captures frames and events within the same pixel array, or alternatively requires beam splitting techniques.

There have been adoptions of frame-based corner detectors to event streams, such as the event-FAST corner detector by Muggler et al. \cite{muggler2017bmcvEfsat} or the event-Harris detector by Vasco et al. \cite{Vasco2016FastEH}. However, those trackers are not robust to changes in motion direction \cite{manderscheid2019speed}. In our work, we circumvent this issue by keeping most information from the past in the form of a template, see section \ref{sec::method}.  

The work by Ignacio et al. \cite{ignacio20193dv} proposes an event-by-event tracking approach which models different hypotheses per tracked feature. While this work tracks features at a very high rate of up to 12500 events per second, it is still formulated in discrete time and thus does not allow for simple derivative calculations of the trajectory, which can be useful in some applications. The approach by Manderscheid et al. \cite{manderscheid2019speed} also performs tracking on an event-by-event basis. They train a random forest which extracts only the corner features from the event stream. The main drawback is that they rely on absolute intensity information during training time.
The work by Zhu et al. \cite{zhu2017icra} first builds feature templates by accumulating the event stream over a short time interval. A batch of new event is then aligned to those templates by probabilistic, soft association. The optical flow is computed as an expectation over all associations and the patch can undergo affine deformations during tracking. 
The work by Seok et al. \cite{Seok2020WACV} is the first approach to formulate event tracking in continuous time. However, adding more knots to the existing B\'ezier curve changes all previous knots, so the feature trajectory has to be formed by concatenating many short B\'ezier curves and is only zero order continuous.

To the best of our knowledge, we are the first to formulate event tracking with the concept of sliding window optimization.


\section{Method} 
\label{sec::method}
We define a B-spline curve $B(t;\Theta,\Delta t)$ which returns a transformation $T_{r,t}$ to transform a 2D-point from current frame at timestamp $t$ to reference frame r. All knots in the spline are denoted by $\Theta$ and the time difference between two knots is denoted by $\Delta t$, which is a pre-determined constant and we will ignore it in our notation later in the paper. If an event $e_n$ is within the region of a patch, it satisfy the condition $B(t_n;\Theta)x_{start} - \textbf{x}_n < R$, where $x_{start}$ is the starting position of the patch and $R$ is the radius of the patch. The position of a warped event $e_n$ is defined as

\begin{equation}
\mathbf{x}_n' = B(t_n;\Theta,\Delta t)\mathbf{x}_n
\end{equation}

\begin{figure}[h]
\centering
\includegraphics[width=130mm]{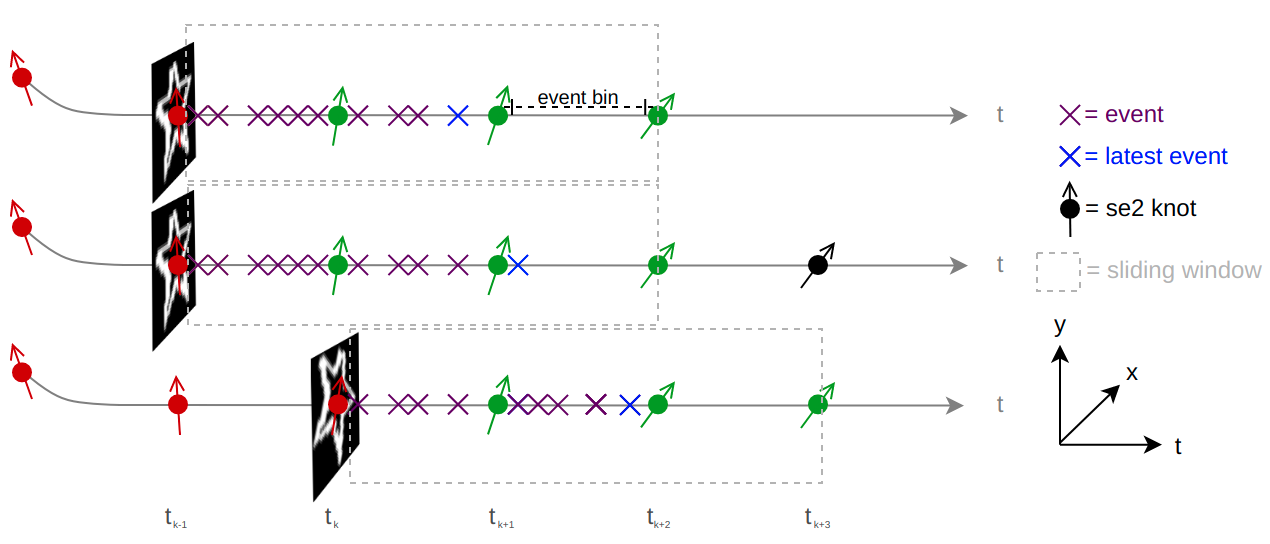}
\caption{Process of updating the sliding window with optimizing 3 knots using SE(2) warping. The arrows on the knots indicate the rotation angles, green knots are the knots inside the sliding window, red knots are fixed knots. The image patches containing the star shape show the History patches. Top: The latest event is in the second last event. Middle: The latest event is in the last event bin, we add an additional knot for the trajectory. Down: After we add a new knot, we update the History patch and the location of the sliding window to fix the problem size.} \label{fig::optWindow}
\end{figure}

We create a patch image by warped back all events in the patch to reference time (here we set the reference time as the starting time of the feature). Since the warped positions of events are not guaranteed to be integers, a bi-linear kernel $k_b$ is used to construct differentiable patches. If event $e_n$ is received at pixel $\mathbf{x}_n$ at time $t_n$, the patch image from events received in the time domain $[t:t']$ is defined as
\begin{equation}
I_{[t:t']}(\mathbf{x}) = \sum_{\{n;t_n \in [t:t']\}} k_b(\mathbf{x},\mathbf{x}_n')
\end{equation}

\begin{equation}
k_b(\mathbf{a},\mathbf{b}) = max(0,1-|a_1-b_1|) \cdot max(0,1-|a_2-b_2|)
\end{equation}

The advantage of using bi-linear kernel instead of using Dirac-delta function is that the patch image is also well-defined in sub-pixel level, which enables us to calculate well-defined $\frac{\partial I}{\partial \mathbf{x'}}$.

In principle, we can optimize a continuous B-spline trajectory by optimizing all knots on the spline and taking all events into account, but this is expensive and inefficient. Inspired by \cite{Seok2020WACV}, we make use of a History patch $H$ which compresses the information of previous knots and events in order to speed up the algorithm. The History patch $H$ is built in recurrent relation. $H$ and the modified patch image $I^*(t)$ at timestamp $t$ are defined as

\begin{equation}
H_{t_{k}}(\mathbf{x};\mathbf{\Theta}) = I_{[t_{k-1}:t_{k})}(\mathbf{x};\mathbf{\Theta}) + \rho H_{t_{k-1}}(\mathbf{x;\mathbf{\Theta}})
\end{equation}

\begin{equation}
I^*(\mathbf{x}, t;\mathbf{\Theta}) = I_{[t_{k}:t)}(\mathbf{x};\mathbf{\Theta}) + H_{t_{k}}(\mathbf{x};\mathbf{\Theta})
\end{equation}
where $t_k$ is the timestamp of knot k, $H_{t_{0}}(\mathbf{x}) = \textbf{0}$ and $0 \leq \rho \leq 1$ is the decaying parameter.

The best B-spline curve is the one which maximize the variance (sharpness) of a patch image from active event with using History patch, see Fig. \ref{fig::optWindow} for an equivalent visualization of the problem using SE(2) warping. The modified variance $\sigma^*$ of the patch at timestamp $t$ is defined as 

\begin{equation}
\label{eqn:modified_var}
\sigma^*(P(t);\mathbf{\Theta}) = \frac{1}{N} \sum_\mathbf{x}  ( I^*(\mathbf{x}, t;\mathbf{\Theta}) - \langle I^*(\mathbf{x}, t;\mathbf{\Theta}\rangle) ^2
\end{equation}

where $N$ is the total number of pixels in the patch, $\mathbf{x}$ is the image coordinate, $\langle I^*(\mathbf{x}, t;\mathbf{\Theta}\rangle$ is the mean value of the modified patch image $I^*$. The work in \cite{gallego2019focus} shows that among 22 possibilities of measuring sharpness in event images, the variance is often a suitable choice.\\

To maximize $I^*$, we need the Jacobian of the variance function w.r.t warping parameters $\mathbf{\Theta}$: 
\begin{equation}
\frac{\partial  I^*(\mathbf{x}, t;\mathbf{\Theta}) }{\partial \mathbf{\Theta}} = \sum_{n=1}^{N_e^i} \frac{\partial k_b(\mathbf{x},\mathbf{x}_n')}{\partial \mathbf{x}_n'} \frac{\partial \mathbf{x}_n'}{\partial B} \frac{ \partial B(t_n;\mathbf{\Theta})}{\partial \mathbf{\Theta}}
\end{equation}

To use SE(2) B-spline as an example, the parameter $\mathbf{\Theta}$ with $N_k$ knots is defined as\\
\begin{equation}
\mathbf{\Theta} = [k_1, \cdots, k_{N_k} ]
\end{equation}
with $k_i = [x_1^i,x_2^i,\theta^i]$

\begin{equation}
\frac{ \partial \mathbf{x}'}{\partial B}
 = -\begin{bmatrix}
R_{r,t} & | & \sigma_x \mathbf{x}' 
\end{bmatrix}_{2\times3}
\end{equation}
with $\sigma_x = \begin{pmatrix} 0 & -1 \\ 1 & 0 \end{pmatrix}$, $R_{r,t}$ is the rotation matrix of the feature relative to its original orientation

\begin{equation}
 \frac{ \partial B(t_n;\mathbf{\Theta})}{\partial k_i}
 = \lambda(t, \Delta t_{knot}) \mathbb{I}_{3\times 3}
\end{equation}
We refer the reader to \cite{sommer19spline} for the deviation of $\lambda(t, \Delta t_{knot})$. The optimal solution $\mathbf{\Theta}^*$ for Equation \ref{eqn:modified_var} is calculated through maximizing $\sigma^*$ by line search.

   

\begin{table}
\caption{Quantitative Comparison of \cite{zhu2017icra} against our method with error threshold 3.}\label{tab::ours-zhu-th3}
\centering
\resizebox{120mm}{!}{%
\begin{tabular}{|
*{1}{>{\centering\arraybackslash}p{0.25\textwidth}}|
*{1}{>{\centering\arraybackslash}p{0.15\textwidth}}||
*{4}{>{\centering\arraybackslash}p{0.17\textwidth}}
*{1}{>{\centering\arraybackslash}p{0.19\textwidth}|}}
\hline
dataset & method & mean relative age & mean age[sec] & mean error[pix] & mean common error[pix] & mean common error for ours/ours$^*$[pix]\\
\hline
\hline
\multirow{3}{*}{shapes\_translation} & ours\;\;        & \textbf{0.27} & \textbf{0.71} & 0.80 & \textbf{0.91} & \textbf{0.95}\\
                                     & ours*       & 0.26 & 0.63 & \textbf{0.79} & \textbf{0.91} & \textbf{0.95}\\
                                     & Zhu et al.  & 0.04 & 0.08 & 2.89 & 2.92 & -\\
\hline
\multirow{3}{*}{shapes\_rotation}    & ours\;\;        & 0.30 & 0.61 & \textbf{0.81} & \textbf{0.93} & \textbf{0.90}\\
                                     & ours*       & \textbf{0.32} & \textbf{0.67} & 0.83 & \textbf{0.93} & \textbf{0.90}\\
                                     & Zhu et al.  & 0.02 & 0.02 & 2.81 & 2.81 & -\\
\hline
\multirow{3}{*}{shapes\_6dof}        & ours\;\;        & \textbf{0.32} & \textbf{1.48} & \textbf{1.05} & \textbf{0.61} & \textbf{1.14}\\
                                     & ours*       & 0.31 & 1.45 & 1.07 & \textbf{0.61} & 1.16\\
                                     & Zhu et al.  & 0.05 & 0.18 & 2.05 & 1.97 & -\\
\hline
\multirow{3}{*}{poster\_translation} & ours\;\;        & 0.47 & 1.43 & \textbf{0.87} & \textbf{0.71} & \textbf{0.88}\\
                                     & ours*       & \textbf{0.48} & \textbf{1.47} & \textbf{0.87} & \textbf{0.71} & \textbf{0.88}\\
                                     & Zhu et al.  & 0.33 & 0.71 & 1.15 & 1.10 & -\\
\hline
\multirow{3}{*}{poster\_rotation}    & ours\;\;        & 0.41 & 1.33 & \textbf{0.79} & \textbf{0.66} & 0.80\\
                                     & ours*       & \textbf{0.45} & \textbf{1.54} & 0.80 & \textbf{0.66} & \textbf{0.79}\\
                                     & Zhu et al.  & 0.20 & 0.51 & 1.51 & 1.40 & -\\
\hline
\multirow{3}{*}{poster\_6dof}        & ours\;\;        & \textbf{0.35} & 2.57 & \textbf{1.05} & \textbf{0.87} & \textbf{1.04}\\
                                     & ours*       & \textbf{0.35} & \textbf{2.61} & \textbf{1.05} & \textbf{0.87} & \textbf{1.04}\\
                                     & Zhu et al.  & 0.25 & 1.37 & 1.32 & 1.28 & -\\
\hline
\multirow{3}{*}{boxes\_translation}  & ours\;\;        & 0.35 & 1.35 & \textbf{0.98} & \textbf{0.88} & \textbf{0.98}\\
                                     & ours*       & \textbf{0.37} & \textbf{1.50} & 1.01 & \textbf{0.88} & 1.00\\
                                     & Zhu et al.  & 0.31 & 0.95 & 1.19 & 0.94 & -\\
\hline
\multirow{3}{*}{boxes\_rotation}     & ours\;\;        & 0.33 & 1.22 & \textbf{0.79} & \textbf{0.68} & 0.81\\
                                     & ours*       & \textbf{0.36} & \textbf{1.41} & 0.80 & \textbf{0.68} & \textbf{0.80}\\
                                     & Zhu et al.  & 0.19 & 0.59 & 1.57 & 1.53 & -\\
\hline
\multirow{3}{*}{boxes\_6dof}         & ours\;\;        & 0.37 & 1.76 & \textbf{1.07} & 0.72 & 1.08\\
                                     & ours*       & \textbf{0.38} & \textbf{1.85} & \textbf{1.07} & \textbf{0.71} & \textbf{1.07}\\
                                     & Zhu et al.  & 0.15 & 0.64 & 1.88 & 1.78 & -\\
\hline
\end{tabular}
}
\end{table}

\section{Experiments}
We evaluate all methods on each dataset with the same pre-selected, evenly-distributed Harris corners. We use a circular patch with diameter $d = 31$, a decaying parameter of $\rho = 0.9$ and track up to 60 features in each experiment. We use third order B-spline and create a new spline knot every 50 milliseconds. If the number of events used in the optimization is small, it may lead to a wrong optimal solution. We tackle this problem by optimizing more knots with more events in this case. In the experiments denoted by \textit{ours*}, we optimize three knots when there are less than $\frac{d^2}{4}$ events in the sliding window, otherwise we optimize only two knots to speed up the run-time. The method of always optimizing two knots is denoted \textit{ours}.


Evaluation is performed on the Event Camera Dataset \cite{mueggler2017event}, which contains recordings from a DAVIS camera. Ground truth feature tracks are computed from frames of the DAVIS camera using the KLT optical flow method \cite{lucas1981iterative}. We compare our methods against Zhu et al. \cite{zhu2017icra}. To allow for a fair comparison against Zhu et al., we use the authors public MATLAB implementation and initialize the tracking with exactly the same feature positions as in our method, disabling the re-detection of new features.

We use four different metrics to do the evaluation. To illustrate the metrics clearly, the error of feature $f_i$ at time $t$ with using method $m$ is denoted by $e^i_m(t)$. The lifetime of feature $f_i$ before the error is larger than threshold $th$ with using method $m$ is denoted by $L_m^{th}(f_i)$. The definition of each metric with error threshold $th$ are:\\

\begin{equation}
\text{mean relative age} = \langle\frac{L_m^{th}(f_i)}{L_{gt}(f_i)}\rangle_i
\end{equation}

\begin{equation}
\text{mean age} = \langle L^{th}_m(f_i)\rangle_i
\end{equation}

\begin{equation}
\text{mean error} = \langle\{ e^i_m(t); \ t\leq L^{th}_m(f_i) \}\rangle_i
\end{equation}

\begin{equation}
\text{mean common error} = \langle\{ e^i_m(t); \ t\leq t_{min}  \}\rangle_i
\end{equation}
where $\langle \dots \rangle_i$ takes average measurement of all features i,  $t_{min}=\min\{(L^{th}_{m_j}(f_i); \  j=1,2,\dots, N_m\}$ is the minimal feature lifetime and $N_m$ is the number of methods we compare. We set the threshold $th$ to 3 pixels in all experiments.

\begin{figure}[h]
\centering
\includegraphics[width=105mm]{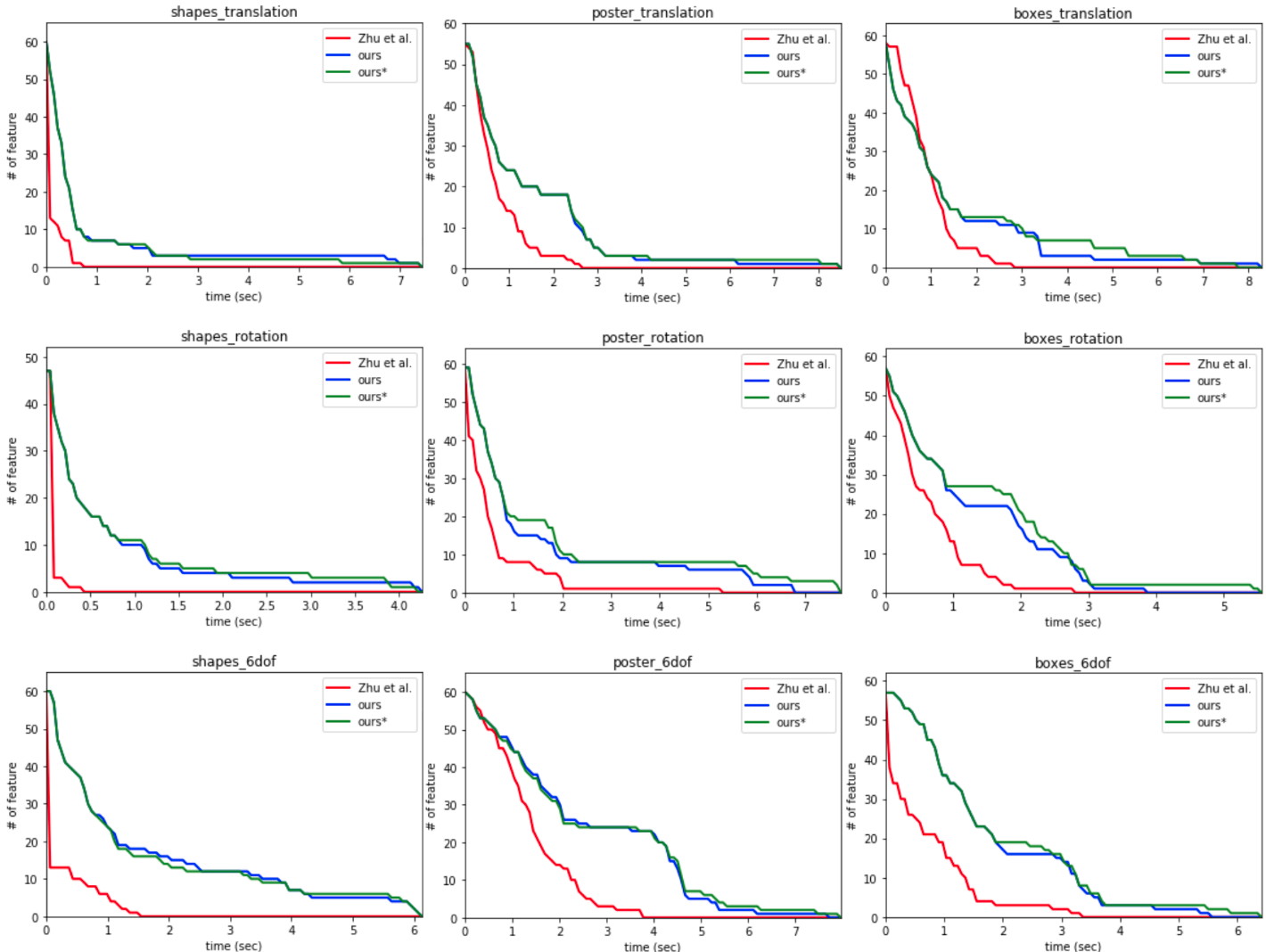}
\caption{Number of feature tracks over time with error threshold 3 for the Event Camera Dataset.} \label{fig::numTracksTime}
\end{figure}

\begin{figure}[h]
\centering
\includegraphics[width=115mm]{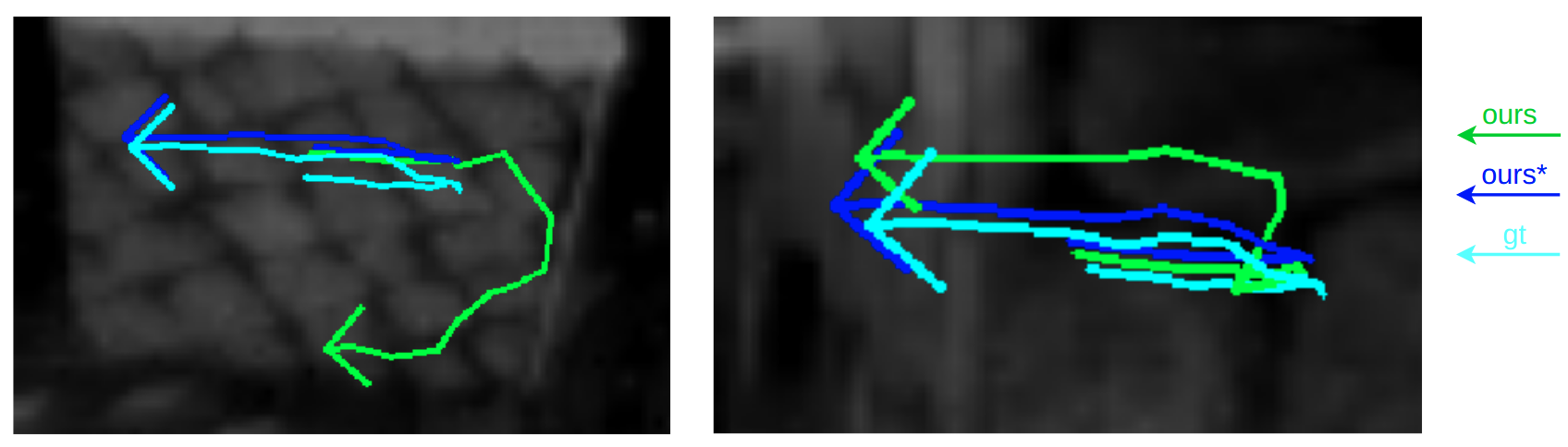}
\caption{Example of the features in Boxes dataset which are improved in method $ours^*$ compared to method $ours$.} \label{fig::example_3knot_better}
\end{figure}

\begin{figure}[h]
\centering
\includegraphics[width=110mm]{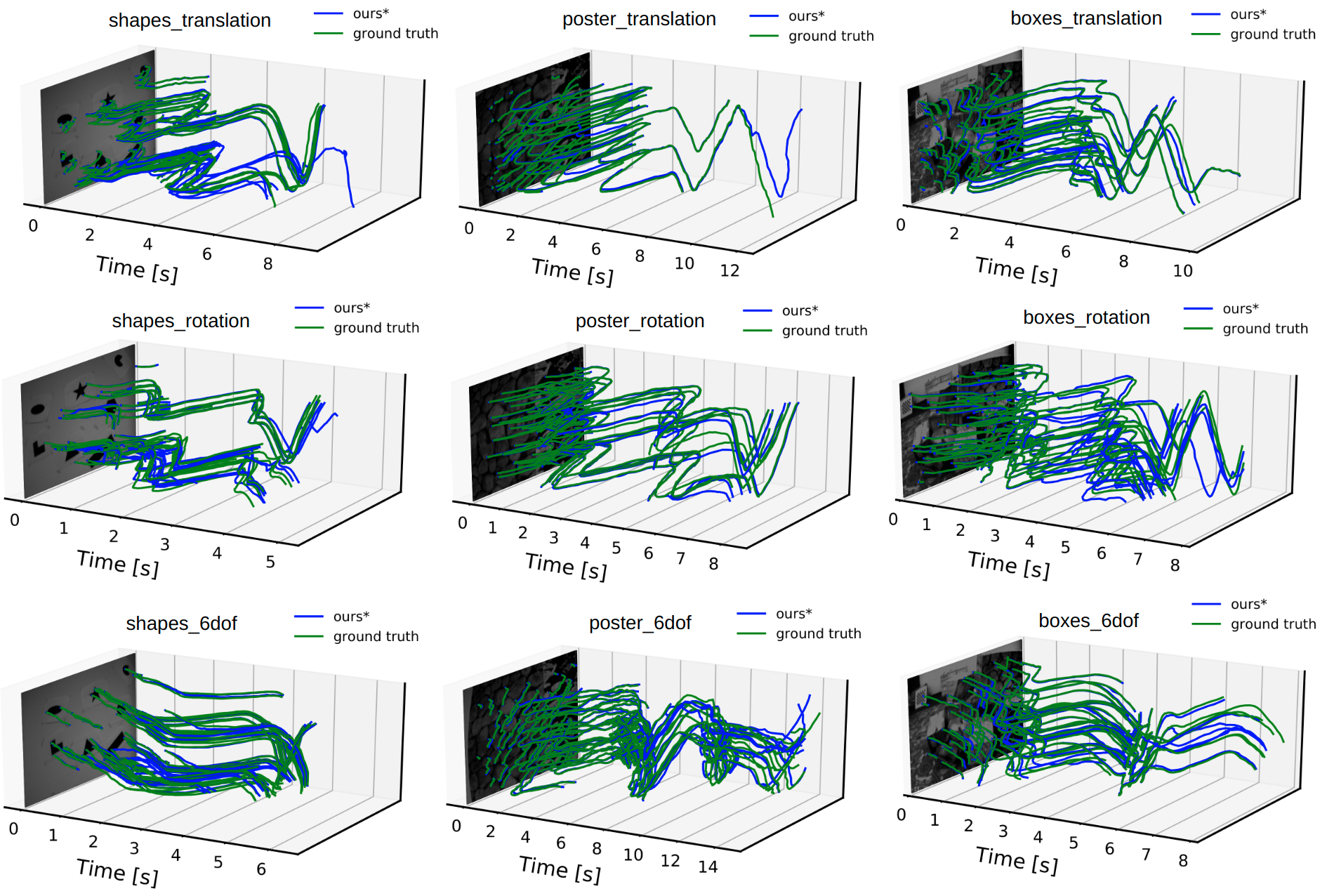}
\caption{Qualitative results of feature tracking for the Event Camera Dataset.} \label{fig::3dplot}
\end{figure}

In Table \ref{tab::ours-zhu-th3}, we compare our method to Zhu's \cite{zhu2017icra} approach. It shows that our method is always better than Zhu. By comparing \textit{ours} and \textit{ours*}, we can see that enlarging the window size during low-events periods can help to improve the mean age around 6\% and the mean common error for ours/ours$^*$ almost remains unchanged. Fig. \ref{fig::example_3knot_better} shows some examples of features which are improved when using method $ours^*$. Fig. \ref{fig::numTracksTime} shows the number of feature tracks over time. The features in $ours*$ last slightly longer than in $ours$. Compared to Zhu, our proposed algorithm tracks more features (with lower error) at almost all instances in time.


Since there is no public implementation of \cite{Seok2020WACV} available, and we are using different initial positions, we can only compare our result qualitatively. In Fig. \ref{fig::3dplot} we show the 3D trajectories which can be used to compare the results to \cite{Seok2020WACV} qualitatively. It shows that our trajectories live longer than theirs.

\section{Conclusion}
In this paper we proposed a novel event tracking algorithm that aligns the event stream with a B-spline curve representation in a sliding window fashion. By using a history patch, the locality of B-splines and a sliding window optimization, our algorithm can track features accurately and for a long time. Our experiments show that the proposed algorithm outperforms the state-of-the-art time-continuous event tracking algorithm. We believe that this method can serve as a basis for event-based video analysis and event-based SLAM.  Future research aims at extending our algorithm to a Sim(2)-formulation, allowing to track features with scale changes.

%
%
%
\bibliographystyle{splncs04}
\bibliography{egbib}

\end{document}